\documentclass[submission,copyright,creativecommons]{eptcs}
\providecommand{\event}{ICLP 2024} 

\usepackage{iftex}

\ifpdf
  \usepackage{underscore}         
  \usepackage[T1]{fontenc}        
\else
  \usepackage{breakurl}           
\fi

\usepackage{hyperref}
\usepackage{amssymb}
\usepackage{subcaption}
\usepackage{tikz-qtree}
\usepackage{enumitem}
\usepackage{fancyvrb}
\usepackage{url}
\usepackage{microtype}
\usepackage{paralist}
\usepackage{xspace}
\usepackage{booktabs} 
\usepackage[T1]{fontenc}

\newcommand{\asplarrow}{\mathrel{\mathrm{:\!\!{-}}}}
\newcommand{\iec}{i.e.,\xspace}
\newcommand{\egc}{e.g.,\xspace}

\newcommand{\ours}{NSGRAPH\xspace}

\newcommand{\CLEGR}{\ensuremath{\mbox{CLEGR}}\xspace}
\newcommand{\ourset}{\ensuremath{\mbox{CLEGR}^V}\xspace}
\newcommand{\CLEGRPLUS}{\ensuremath{\mbox{CLEGR}^+}\xspace}
\newcommand{\CLEGRHUMAN}{\ensuremath{\mbox{CLEGR-Human}}\xspace}

\newtheorem{example}{Example}

\title{Visual Graph Question Answering with ASP and \\ LLMs for Language Parsing\thanks{This work was partially funded from the Bosch Center for AI. Code and data can be found at \url{https://github.com/pudumagico/NSGRAPH}.}}

\author{ {Jakob Johannes} {Bauer$^1$}, {Thomas} {Eiter$^2$}, {Nelson} {Higuera Ruiz$^2$}, {Johannes} {Oetsch$^3$}
\institute{$^1$ ETH Zürich, Rämistrasse 101, 8092 Zürich, Switzerland \\
$^2$ Vienna University of Technology \textup{(}TU Wien\textup{)}, Favoritenstrasse 9--11, Vienna, 1040, Austria \\
$^3$ Jönköping University, Gjuterigatan 5, 551$\,$11 Jönköping, Sweden \\
{\tt bjohannes@ethz.ch}, {\{{\tt thomas.eiter,nelson.ruiz}\}@tuwien.ac.at}, {\tt johannes.oetsch@ju.se}
}}

\def\titlerunning{Visual Graph Question Answering with ASP and LLMs for Language Parsing}
\def\authorrunning{Jakob Johannes Bauer, Thomas Eiter, Nelson Higuera, Johannes Oetsch}
\def\copyrightholders{J.J. Bauer, Th. Eiter, N. Higuera \& J. Oetsch}

\begin{document}
\maketitle
\begin{abstract}
Visual Question Answering (VQA) is a challenging problem that requires to process multimodal input. Answer-Set Programming (ASP) has shown great potential in this regard to add interpretability and explainability to modular VQA architectures. In this work, we address the problem of how to integrate ASP with modules for vision and natural language processing to solve a new and demanding VQA variant that is concerned with images of graphs (not graphs in symbolic form). Images containing graph-based structures are an ubiquitous and popular form of visualisation. Here, we deal with the particular problem of graphs inspired by transit networks, and we introduce a novel dataset that amends an existing one by adding images of graphs that resemble metro lines. Our modular neuro-symbolic approach combines optical graph recognition for graph parsing, a pretrained optical character recognition neural network for parsing labels, Large Language Models (LLMs) for language processing, and ASP for reasoning. This method serves as a first baseline and achieves an overall average accuracy of 73\% on the dataset. Our evaluation provides further evidence of the potential of modular neuro-symbolic systems, in particular with pretrained models that do not involve any further training and logic programming for reasoning, to solve complex VQA tasks.
\end{abstract}


\section{Introduction}\label{sec:intro}

\begin{figure}[t]
     \centering \footnotesize
     \begin{subfigure}[b]{0.25\textwidth}
         \centering
         \includegraphics[width=\textwidth]{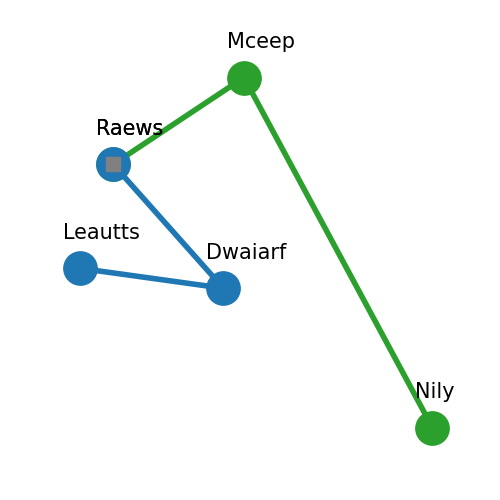}
         \caption{}
         \label{subfig:graph}
     \end{subfigure}
     \hfill
     \begin{subfigure}[b]{0.25\textwidth}
         \centering
        ``How many stations are between Leauts and Nily?''\\
        \vspace{5pt}
        \resizebox{\columnwidth}{!}{%
        \begin{tikzpicture}[scale=1,level distance=35pt]
        \Tree [ .\textit{Substract}() 
                [ .\textit{Count}() 
                    [ .\textit{ShortestPath}() 
                        [ .\textit{Station}(Leauts)  ] 
                        [  .\textit{Station}(Nily) ] ] ] 
                [ .\textit{Int}(2) ] ]
        \end{tikzpicture}
        }
        \caption{}
        \label{subfig:question}
     \end{subfigure}
     \hfill
     \begin{subfigure}[b]{0.25\textwidth}
        \setlist{nolistsep}
        \begin{compactitem}
            \item nodes:
            \begin{compactitem}
                \item[] name: leautts\\
                size: tiny\\
                \ldots
            \end{compactitem}
            \item edges:
            \begin{compactitem}
                \item[] station1: raows\\
                        station2: dwaiarf\\
                \ldots
            \end{compactitem}
            \item lines:
            \begin{compactitem}
                \item[] built: 90s\\
                \ldots
            \end{compactitem}
        \end{compactitem}
        \caption{}
         \label{subfig:bgknowledge}
     \end{subfigure}
    \caption{A \CLEGR instance: metro graph with two lines, question with functional representation, and additional information. The task is to answer the question  using the information provided.}
        
        \label{fig:gqa-instance}
\end{figure}

\emph{Visual Question Answering} (VQA)~\cite{antol15} is concerned
with inferring the correct answer to a natural language question in the presence of some visual input, such as an image or video, which typically involves processing multimodal input. VQA enables applications in, \egc medicine, assistance for blind people, surveillance, and education~\cite{barra2021vqaapps}.

\emph{Answer-Set Programming} (ASP)~\cite{brewka2011asp} has shown great potential to add interpretability and explainability to modular VQA architectures in this context. 
As a knowledge representation and reasoning formalism with an intuitive modelling language, it can be used to describe how to infer answers from symbolic input provided by subordinate modules in a clear and transparent way~\cite{RileyS19,BasuSG20,eiterHOP22,EiterGHO23}. Another strength is that uncertainties from the underlying modules can be expressed using disjunctions (or choice rules), and we are not limited to inferring one answer, but several plausible ones in a nondeterministic manner~\cite{yang2020neurasp}.
Furthermore, using ASP in the VQA context is 
beneficial for explanation finding, as we have demonstrated in recent work~\cite{EiterGHO23}.

In this work, we address the problem of how to integrate ASP with modules for vision and natural language processing to solve a new and demanding VQA variant that is concerned with images of graphs (not graphs in symbolic form). Visual representations of structures based on graphs are a popular and ubiquitous form of presenting information.
It is almost surprising that VQA tasks where the visual input contains a graph have, to the best of our knowledge, not been considered so far.

We deal with the particular problem of graphs that resemble transit networks, and we introduce a respective dataset. It is based on the existing  \CLEGR\ dataset~\cite{clevr-grah} that comes with a generator for synthetically producing vertex-labelled graphs that are inspired by metro networks. Additional structured information about stations and lines, \egc how large a station is, whether it is accessible to disabled people, when the line was constructed, etc., is provided as background. The task is to answer natural language questions concerning such graphs. 
For example, a question may ask for the shortest path between two stations while avoiding those that have a particular property. An illustration of a graph and a question is shown in Fig.~\ref{fig:gqa-instance}.

While purely symbolic methods suffice to solve the original \CLEGR dataset with ease (we present one in this paper), we consider the more challenging problem of taking images of the graphs instead of their symbolic representations as input; an example is given in Fig.~\ref{subfig:graph}. 
For the questions, we only consider those that can be answered with information that can be found in the image.
The challenges to solve this VGQA dataset, which we call \ourset, are threefold: (i) we have to parse the graph to identify nodes and edges, (ii) we have to read and understand the labels and associate them with nodes of the graph, and (iii) we have to understand the question and reason over the information extracted from the image to answer it accordingly. 

Our solution takes the form of a modular (\iec loosely coupled) neuro-symbolic model 
that combines the use of \emph{optical graph recognition} (OGR)~\cite{auer2013ogr} for graph parsing, a pretrained \emph{optical character recognition} (OCR) neural network~\cite{abin2018ocr}  for parsing node labels, and, as mentioned above,  ASP for reasoning. 
It operates in the following manner: 
\begin{compactenum}

\item 
first, we use the OGR tool to parse the graph image into an abstract representation, structuring the information as sets of nodes and edges;

\item 
we use the OCR algorithm to obtain the text labels and associate them to the closest node;

\item 
then, we parse the natural language question; 

\item 
finally, we use an encoding of the semantics of the question as a logic program  which is, combined with the graph and the question in symbolic form, used to obtain the answer to the question with the help of an ASP solver.
\end{compactenum}
\noindent
This method serves as a first baseline and achieves an average accuracy of 73\% on \ourset.

We consider two methods to parse the natural language questions.
The first one is to use regular expressions which are sufficient to parse the particular questions of the dataset. The second method uses Large Language Models (LLMs) based on the transformer architecture~\cite{vaswani17} to obtain a more robust solution that also generalises well to variants of questions that are not part of the dataset. Our approach to using LLMs follows related work~\cite{rajasekharan23} and relies on prompting an LLM to extract relevant ASP predicates from the question. We evaluated this approach on questions based on \CLEGR and new questions obtained from a questionnaire.

The contribution of this paper is thus threefold:
\begin{compactenum}[(i)]
\item
we demonstrate how ASP can be used as part of a modular VQA architecture able to tackle a challenging new problem concerned with images of graphs;

\item we introduce a new dataset to benchmark systems for VQA on images of graphs and evaluate our approach on it to create a first baseline; and

\item 
we evaluate various {LLMs for question parsing} to create a robust interface to the ASP encoding.
\end{compactenum}

This work provides further evidence of the potential of modular neuro-symbolic systems, in particular with pretrained models and logic programming for reasoning, for solving complex VQA tasks. That our system does not require any training related to a particular set of examples---hence solving the dataset in a \emph{zero-shot manner}---is a practical feature that hints to what may become customary as large pre-trained models are more than ever available for public use.

\section{Visual Question Answering on Graphs}\label{sec:vgqa}

\begin{figure}[t]
     \centering
     \begin{subfigure}[b]{0.25\textwidth}
         \centering
         \includegraphics[width=\textwidth]{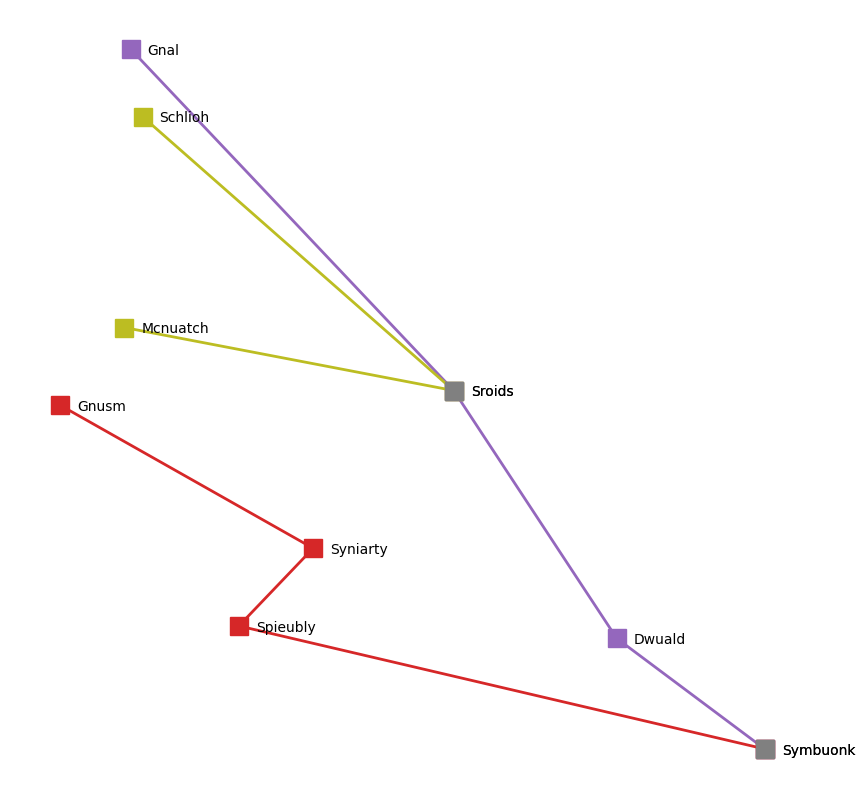}
         \caption{}
         \label{subfig:small}
     \end{subfigure}
     \hfill
     \begin{subfigure}[b]{0.25\textwidth}
         \centering
         \includegraphics[width=\textwidth]{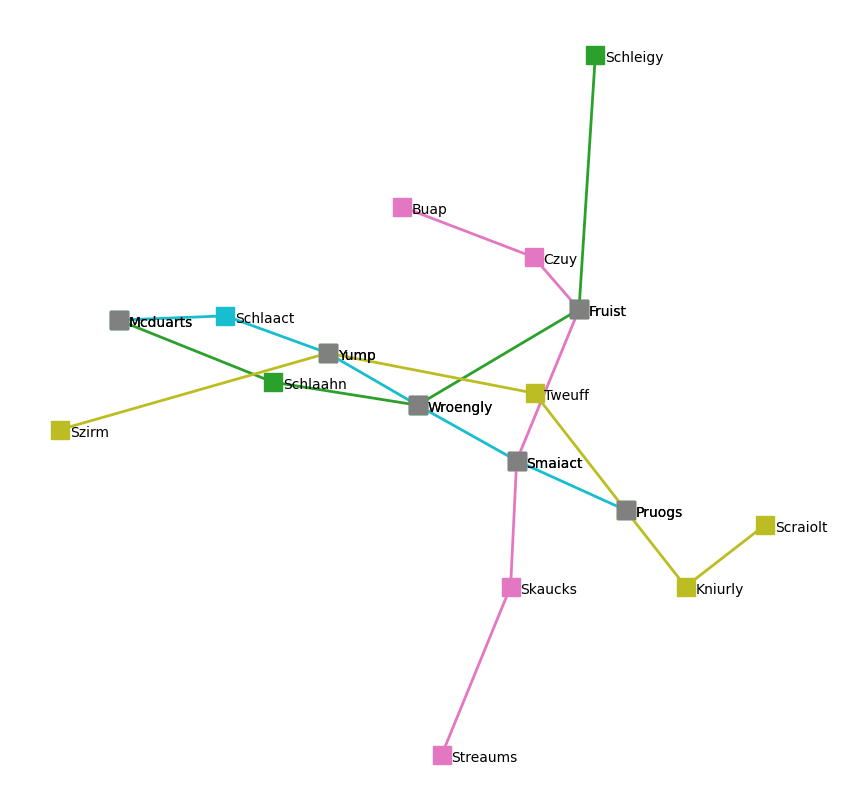}
         \caption{}
         \label{subfig:medium}
     \end{subfigure}
     \hfill
     \begin{subfigure}[b]{0.25\textwidth}
         \centering
         \includegraphics[width=\textwidth]{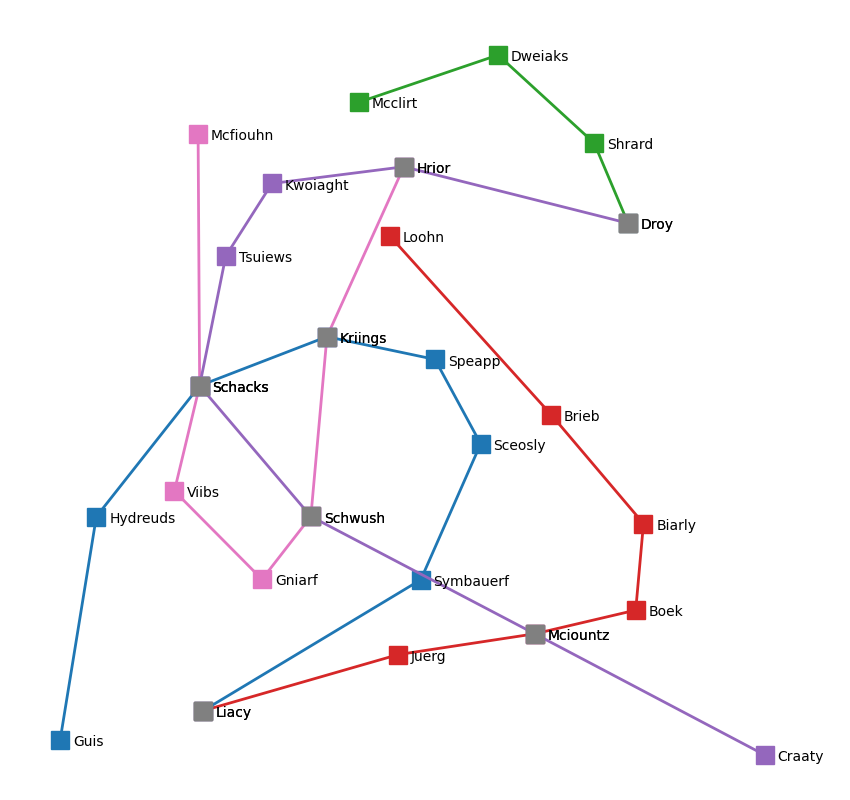}
         \caption{}
         \label{subfig:large}
     \end{subfigure}
     \caption{Examples of graphs of size small (\ref{subfig:small}), medium (\ref{subfig:medium}), and large (\ref{subfig:large}).}
     \label{fig:graph-sizes}
\end{figure}


\paragraph{Graph Question Answering (GQA) }
is the task of answering a natural language question for a given graph in symbolic form. The graph consists of nodes and edges, but further attributes may be specified in addition.
%
A specific GQA dataset is \CLEGR~\cite{clevr-grah}, which is concerned with graph structures that
resemble transit networks like metro lines. 
Its questions are ones that are typically asked about transit like ``How many stops are between {\bf X} and {\bf Y}?''.
The dataset is synthetic and comes with a generator for producing instances of varying complexity. 

Graphs come in the form of a YAML file containing records about attributes of the stations and lines. 
Each station has a name, a size, a type of architecture, a level of cleanliness, potentially disabled access, potentially rail access, and a type of music played.  Stations can be described as relations over the aforementioned attributes. 
Edges connect stations but additionally have a colour, a line ID, and a line name. 
For lines, besides name and ID we have a construction year, a colour, and optional presence of air conditioning.

\begin{example}\label{ex:1}
Examples of questions from the dataset are:
\begin{compactitem}
\item 
Describe \{Station\} station's architectural style.
\item 
How many stations are between \{Station\} and \{Station\}?
\item 
Which \{Architecture\} station is adjacent to \{Station\}?
\item 
How many stations playing \{Music\} does \{Line\} pass through?
\item 
Which line has the most \{Architecture\} stations?
\end{compactitem}
\end{example}
For a full list of the questions, we refer the reader to the online repository of the dataset~\cite{clevr-grah}. 
The answer to each question is of Boolean type, a number, a list, or a categorical answer.
The questions in the dataset can be represented by functional programs, which allows us to decompose them into smaller and semantically less complex components.
Figure~\ref{fig:gqa-instance} illustrates an example from the data set \CLEGR that includes such a functional program: it consists of primitive operations organised as a tree that is recursively evaluated to obtain an answer.  

\paragraph{Visual Graph Question Answering.}
Solving instances of the \CLEGR dataset is not much of a challenge since all information is given in symbolic form,  and we present a respective method later. But what if the graph is not available or given in symbolic form, but just as an image, as is commonly the case?
We define \emph{Visual Graph Question Answering} (VGQA) as a GQA task where the input is a natural language question on a graph depicted in an image. 

\paragraph{The new VGQA dataset.}
We can in fact derive a challenging VGQA dataset from \CLEGR by generating images of the transit graphs. To this end, we used the generator of the \CLEGR dataset that can also produce images of the symbolic graphs.
Each image shows stations, their names as labels in their proximity, and lines in different colours that connect them; an example is given in Fig.~\ref{subfig:graph}.
For the VGQA task, we drop all further symbolic information and consider only the subset of questions that can be answered with information from the graph image.

We call the resulting dataset \ourset: 
%
it consists of graphs that fall into three categories: small (3 lines and at most 4 stations per line), medium (4 and at most 6 stations per line), and large (5 lines and at most 8 stations per line). We generate 100 graphs of each size accompanied by 10 questions per graph, with a median of 10 nodes and 8 edges for small graphs, 15 nodes and 15 edges for medium graphs, and 24 nodes with 26 edges for large ones. Figure~\ref{fig:graph-sizes} shows three graphs, one of each size. Although large metro networks will typically involve more stations than our graphs, those stations are typically arranged linearly on the lines which does not add to the complexity of the graph structure itself but can lead to cluttering.

\section{Our Neuro-Symbolic Framework for VQA on Graphs}\label{sec:framework}

\begin{figure}[t]
    \centering
    \includegraphics[width=.9\textwidth]{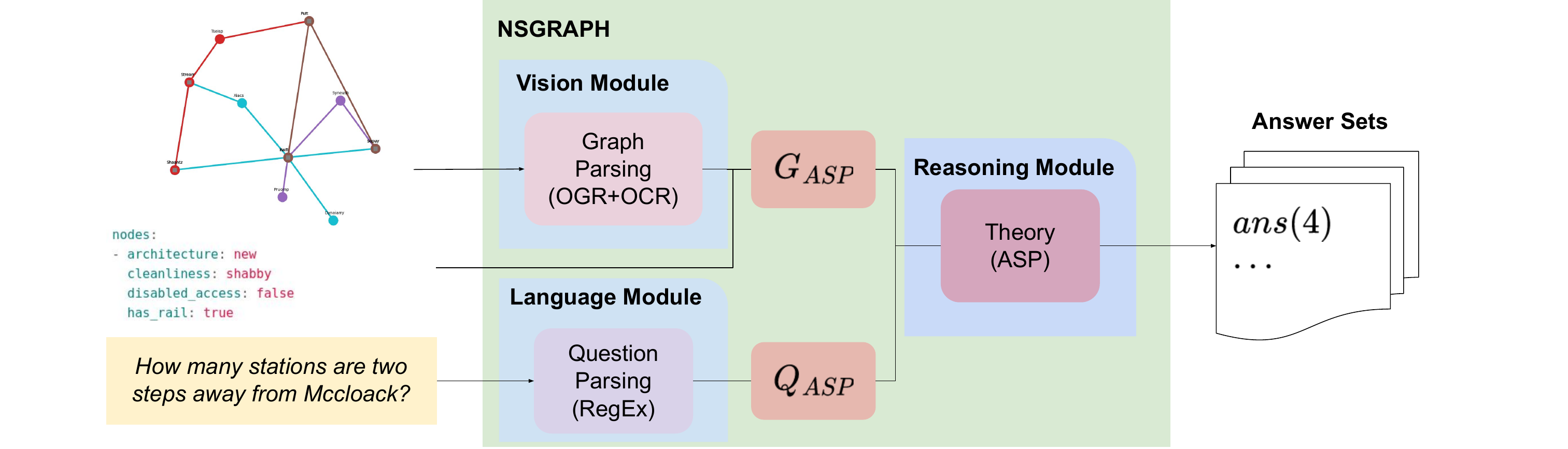}
    \caption{\ours system overview. The input is either an image of a graph or its symbolic description. The answer is generated by combining neural and symbolic methods.}
    \label{fig:sys}
\end{figure}

Our solution to the VGQA task, which we call \ours,
is a modular neuro-symbolic system, 
whose modules are the typical ones for VQA, viz.\  a visual module, a language module, and a reasoning module, which we realise to fit the VQGA setting. Figure~\ref{fig:sys} 
illustrates the data flow of the inference process in \ours.

\subsection{Visual Module} 

The visual model is used for graph parsing, which consists of two subtasks: (i) detection of nodes and edges, and (ii) detection of labels, \iec station names.

We employ an optical graph recognition (OGR) system for the first subtask. In particular, 
we use a publicly available OGR script~\cite{ogr} that implements the approach 
due to Auer et al.~\cite{auer2013ogr}.
The script takes an image as input and outputs the pixel coordinates of each detected node
plus an adjacency matrix that contains the detected edges.

For the second subtask of detecting labels, we use an optical character recognition (OCR)
system, namely, we use a pretrained neural network
called EasyOCR~\cite{ocr} to obtain and structure the information contained in the graph image.
The algorithm takes an image as input and produces the labels as strings together with their coordinates in pixels.
We then connect the detected labels to the closest node found by the OGR system. Thereby, we obtain an abstract representation of the graph image as relations.

\subsection{Language Module} 

The purpose of the language module is to parse the natural language question.
It is written in Python and uses regular expressions to capture the variables in each type of question. There are in general 35 different question templates in \CLEGR, some of which were shown in Example~\ref{ex:1}. They can be used to produce a question instances by replacing variables with names or attributes of stations, lines, or connections.
\begin{example}
For illustration, the question template ``How many stations are on the shortest path between $S_1$ and $S_2$?'' may be instantiated by replacing $S_1$ and $S_2$ with station names that appear in the graph.
We use regular expressions to capture those variables and translate the natural language question into a functional program, essentially a tree of operations, for that question.
Continuing our example, we translate the template described above into the program 

{ \small
\begin{verbatim}
end(3). countNodesBetween(2). shortestPath(1). 
station(0,S1). station(0,S2).
\end{verbatim}
}

\noindent
where the the first numerical argument of each predicate imposes the order of execution of the associated operation and links the input of one operation to the output of the previous one.
We can interpret this functional program as follows: the input to the shortest-path operation is two station names \verb!S1! and \verb!S2!. Its outputs are the stations on the shortest path between \verb!S1! and \verb!S2! which are counted in the next step. The predicate \verb!end! represents the end of the computation to yield this number as the answer to the question.
\end{example}


\begin{table}[t]
\small
\centering
\caption{ASP questions encodings for the twelve types of questions.}
\label{tab:predicatesUsed}
{\begin{tabular}{@{\extracolsep{\fill}}l@{\qquad}l} 
\toprule
 ASP Facts  & Question \\
\midrule
\begin{minipage}[t]{0.42\textwidth}\raggedright
end(3). countNodesBetween(2).
shortestPath(1). station(0,\{\}). station(0,\{\})
\end{minipage} &
\begin{minipage}[t]{0.5\textwidth}\raggedright
How many stations are between ([a-zA-Z]+) and ([a-zA-Z]+)?
\end{minipage} \\
\midrule
\begin{minipage}[t]{0.42\textwidth}\raggedright
end(2). withinHops(1, 2). station(0,\{\})
\end{minipage} &
\begin{minipage}[t]{0.5\textwidth}\raggedright
How many other stations are two stops or closer to ([a-zA-Z]+)?
\end{minipage} \\
\midrule
 \begin{minipage}[t]{0.42\textwidth}\raggedright
end(2). paths(1). station(0,\{\}). station(0,\{\})
\end{minipage} &
\begin{minipage}[t]{0.5\textwidth}\raggedright
How many distinct routes are there between ([a-zA-Z]+) and ([a-zA-Z]+)?
\end{minipage} \\
\midrule
 \begin{minipage}[t]{0.42\textwidth}\raggedright
end(2). cycle(1). station(0,\{\})
\end{minipage} &
\begin{minipage}[t]{0.5\textwidth}\raggedright
Is ([a-zA-Z]+) part of a cycle?
\end{minipage} \\
\midrule
 \begin{minipage}[t]{0.42\textwidth}\raggedright
end(2). adjacent(1). station(0,\{\}). station(0,\{\})
\end{minipage} &
\begin{minipage}[t]{0.5\textwidth}\raggedright
Are ([a-zA-Z]+) and ([a-zA-Z]+) adjacent?
\end{minipage} \\
\midrule
 \begin{minipage}[t]{0.42\textwidth}\raggedright
end(2). adjacentTo(1). station(0,\{\}).station(0,\{\})
\end{minipage} &
\begin{minipage}[t]{0.5\textwidth}\raggedright
Which station is adjacent to ([a-zA-Z]+) and ([a-zA-Z]+)?
\end{minipage} \\
\midrule
 \begin{minipage}[t]{0.42\textwidth}\raggedright
end(2). commonStation(1). station(0,\{\}). station(0,\{\})
\end{minipage} &
\begin{minipage}[t]{0.5\textwidth}\raggedright
Are ([a-zA-Z]+) and ([a-zA-Z]+) connected by the same station?
\end{minipage} \\
\midrule
 \begin{minipage}[t]{0.42\textwidth}\raggedright
end(2). exist(1). station(0,\{\})
\end{minipage} &
\begin{minipage}[t]{0.5\textwidth}\raggedright
Is there a station called ([a-zA-Z0-9]+)?
\end{minipage} \\
\midrule
 \begin{minipage}[t]{0.42\textwidth}\raggedright
end(2). linesOnNames(1). station(0,\{\})
\end{minipage} &
\begin{minipage}[t]{0.5\textwidth}\raggedright
Which lines is ([a-zA-Z]+) on?
\end{minipage} \\
\midrule
 \begin{minipage}[t]{0.42\textwidth}\raggedright
end(2). linesOnCount(1). station(0,\{\})
\end{minipage} &
\begin{minipage}[t]{0.5\textwidth}\raggedright
How many lines is ([a-zA-Z]+) on?
\end{minipage} \\
\hline
 \begin{minipage}[t]{0.42\textwidth}\raggedright
end(2). sameLine(1). station(0,\{\}). station(0,\{\})
\end{minipage} &
\begin{minipage}[t]{0.5\textwidth}\raggedright
Are ([a-zA-Z]+) and ([a-zA-Z]+) on the same line?
\end{minipage} \\
\midrule
 \begin{minipage}[t]{0.42\textwidth}\raggedright
end(2). stations(1). line(0,\{\})
\end{minipage} &
\begin{minipage}[t]{0.5\textwidth}\raggedright
Which stations does ([a-zA-Z]+) pass through?
\end{minipage} \\
\bottomrule
\end{tabular}}
\end{table}

All considered question types and their ASP question encodings are summarised in Table~\ref{tab:predicatesUsed}.
Although this approach works well for all the questions in \CLEGR, its ability to generalise to new types of questions is obviously limited; as a remedy, we discuss LLMs as an alternative to realise the language module in Section~\ref{sec:LLM}.

\subsection{Reasoning Module} 

The third module consists of an ASP program that implements the semantics of the operations from the functional program of the question. 
Before we explain this reasoning component, 
we briefly review the basics of ASP.

\paragraph{Answer-Set Programming.}

ASP~\cite{brewka2011asp,gebser2012}
is a declarative logic-based approach to combinatorial search and optimisation with roots in knowledge representation and reasoning. It offers a simple modelling language and efficient solvers\footnote{See, for example, \url{www.potassco.org} or \url{www.dlvsystem.com}.}.
In ASP, the search space and properties of problem solutions are
described by means of a logic program such that its models, called \emph{answer sets}, encode the problem solutions.

An ASP program is a set of rules of the form
{
$a_1 \mid \cdots \mid a_m \asplarrow\ b_1,
\ldots,\ b_n,\ not\ c_1, \ldots,\ \mathit{not}\ c_n$, 
}
\noindent
where all $a_i$, $b_j$, $c_k$ are first-order literals and $\mathit{not}$
is \emph{default negation}.
The set of atoms left of $\asplarrow$ is the head of the rule, while the atoms to the right
form the body.
Intuitively, whenever all $b_j$ are true and there is no evidence for any $c_l$, then at least some $a_i$ must be true. 
The semantics of an ASP programs is given by its answer sets, which are consistent sets of variable-free (ground) literals that satisfy all rules and fulfil a minimality condition~\cite{GelfondL91}.

A rule with an empty body and a single head atom without variables is a \emph{fact} and is always true.
A rule with an empty head is a \emph{constraint} and is used to exclude models that satisfy the body.

ASP provides further language constructs like choice rules, aggregates, and weak (also called soft) constraints, whose violation should only be avoided. For a comprehensive coverage of the ASP language and its semantics, we refer to the language standard~\cite{calimeri2020aspcore}.

\paragraph{Question Encoding.} 

The symbolic representations obtained from the language and visual modules are first translated 
into ASP facts; we refer to them as $G_{ASP}$ and $Q_{ASP}$ in Fig.\ref{fig:sys}, respectively. The functional program from a question (as introduced above) is already in a fact format. The graph is translated into binary atoms \verb!edge/2! and unary atoms \verb!station/1! as well.
These facts combined with an ASP program that encodes the semantics of all \CLEGR question templates can be used to compute the answer with an ASP solver.

\begin{example}
Here is an excerpt of the ASP program that represents the functional program from above:
{ \small
\begin{verbatim}
end(3). countNodesBetween(2). shortestPath(1). 
station(0,s). station(0,t).
\end{verbatim}
}
These facts, together with ones for edges and nodes, serve as input to the ASP encoding for computing the answer as they only appear in rule bodies:

\begin{center}
{ \small
\begin{verbatim}
sp(T,S1,S2) :- shortestPath(T), station(T-1,S1), 
                                station(T-1,S2), S1<S2'.
{ in_path(T,S1,S2) } :- edge(S1,S2), shortestPath(T).
reach(T,S1,S2) :- in_path(T,S1,S2).
reach(T,S1,S3) :- reach(T,S1,S2), reach(T,S2,S3).
:- sp(T,S1,S2), not reach(T,S1,S2).

cost(T,C) :- C = #count {S1,S2: in_path(T,S1,S2)}, shortestPath(T).
:~ cost(T,C). [C,T]

countedNodes(T,C-1) :- countNodesBetween(T), 
                       shortestPath(T-1), cost(T-1,C).
ans(N) :- end(T), countedNodes(T,N).
\end{verbatim}
}
\end{center}

The first rule expresses that if we see \verb!shortestPath(T)! in the input, then 
we have to compute the shortest path between station \verb!S1! and \verb!S2!.
This path is produced by the next rule which non-deterministically decides for every edge if this edge is part of the path. The following two rules jointly define the transitive closure of this path relation, and the constraint afterwards enforces that station \verb!S1! is reachable from \verb!S2! on that path.
We use a weak constraint to minimise the number of edges that are selected and thus enforce that we indeed get a shortest path. The number of edges is calculated using an aggregate expression to count. Finally, the penultimate rule calculates the number of stations on the shortest path, as it takes as input the nodes that came out of the shortest path from the previous step and counts them, and the last rule defines the answer to the question as that number. 
The complete encoding is part of the online repository of this project (\url{https://github.com/pudumagico/NSGRAPH}). 
\end{example}

%


\subsection{Evaluation of \ours  on \ourset}\label{sec:expr} 

\begin{table}[t]
    \centering
    \caption{Accuracy of \ours on \ourset for small, medium, and large sized graphs. For OCR+GT, we replaced the OGR input with its symbolic ground truth. Likewise, 
    we use the ground truth for OCR for OGR+GT, and Full GT stands for ground truth only. We also report the total time for image parsing, resp. ASP reasoning, in seconds.} 
    \label{tab:results}
    \small
    \begin{tabular}{@{\extracolsep{\fill}}l r r r r  r r}
    \toprule
         Graph Size       &  \ours &OCR+GT & OGR+GT & Full GT & parsing (s)& reasoning (s) \\
    \midrule
         Small          & 80.9\% &90.2\%  & 83.1\% & 100\% &  923& 2\\
         Medium         & 71.0\% &85.2\%  & 72.7\% & 100\% & 1359& 3 \\
         Large        & 67.2\% &83.8\%  & 70.5\% & 100\%   & 2208& 5  \\
         \midrule
         Overall     & 73.0\%   &86.4\%  & 75.4\% & 100\% & 	4490 & 10\\
    \bottomrule
    \end{tabular}
    \vspace{7pt}
\end{table}

\ours achieves 100\% on the original GQA task, \iec with graphs in symbolic form as input and with the complete set of questions. Here, the symbolic input is translated directly into ASP facts without the need to parse an image.

We summarise the results for the more challenging VGQA task on \ourset in Table~\ref{tab:results}.\footnote{%
We ran the experiments on a computer with 32GB RAM, 12th Gen Intel Core i7-12700K, and a NVIDIA GeForce RTX 3080 Ti, and we used clingo (v.~5.6.2)~\cite{gebser2016theory} as ASP solver. 
}
The task becomes more difficult with increasing size of the graphs, but still an overall accuracy of $73\%$ is achieved. As we also consider settings where we replace the OCR, resp.\ the OGR module, with the ground truth as input, we are able to pinpoint the OGR as the main reason for wrong answers.  
The average run time to answer a question was $0.924\,s$ for small graphs, $1.36\,s$ for medium graphs, and $2.21\,s$ for large graphs.
\ours is the first baseline for this VGQA dataset and further improvements a certainly possible, \egc stronger OGR systems could be used.

\section{Semantic Parsing with LLMs}\label{sec:LLM}

LLMs like GPT-4~\cite{openai2023gpt4} are deep neural networks based on the transformer architecture~\cite{vaswani17} with billions of parameters that are trained on a vast amount of data to learn to predict the next token for a given text prompt. (A token is a sequences of textual characters like words or parts of words).
Their capabilities for natural language processing are impressive. LLMs are typically instructed via text prompts to perform a certain task such as answering a question or translating a text, but they can also be used for semantic parsing a text into a formal representation suitable for further processing.

 In this section, we outline and evaluate an approach to use LLMs to realise the language module of \ours in a more robust way than by using regular expressions. First, we outline the general method of prompting LLMs to extract ASP predicates from questions. Afterwards, we evaluated this method for different LLMs,  including state-of-the-art API-based ones but also open-source models that are free and can be locally installed.

\subsection{Prompt Engineering}   

A particularly useful feature of LLMs is that the user can instruct them for a task by providing a few examples as part of the input prompt without the need to retrain the model on task-specific data;  a property of LLMs commonly referred to as \emph{in-context learning}. 

Our approach uses in-context learning to instruct the LLM to extract the ASP atoms needed to solve the reasoning task from a question. This idea is inspired by recent work on LLMs for language understanding~\cite{rajasekharan23}. To obtain an answer to a question Q, we 
\begin{compactenum}[(i)]
\item create a prompt P(Q) that contains the question Q along with additional instructions and examples for ASP question encodings,  
\item pass P(Q) as input to an LLM and extract the ASP question encoding from the answer, and
\item use extracted ASP facts together with the ASP rules described in the previous section to derive the answer.
\end{compactenum}

The prompt P(Q) starts with a general pre-prompt that sets the stage for the task: 
{ \small
\begin{verbatim}
You are now a Question Parser that translates natural language 
questions into ASP ground truths about different stations. 
Output only the ground truths and nothing else. The stations to 
be selected from are arbitrary.
\end{verbatim}
}

Afterwards, we provide a number of examples that illustrate what is expected from the LLM. In particular, we used at least one not more than three examplesfor each type of question in the dataset to not exceed context limits.
This amounts to 36 in-context examples in total.  

\begin{example}
For space reasons, we show here just the beginning of an example prompt:

{ \small
\begin{verbatim}
I now provide you with some examples on how to parse Questions:

Q: ``How many stations are between Inzersdorf and Mainstation?''
A: end(3).countNodesBetween(2).shortestPath(1).
station(0,``Inzersdorf'').station(0,``Mainstation'').

Q: ``What is the amount of stations between Station A and 
Station B?''
A: end(3).countNodesBetween(2).shortestPath(1).
station(0,``Station A'').station(0,``Station B'').
...
\end{verbatim}
}
\end{example}
Finally, the prompt contains the questions that should be answered:
{ \small
\begin{verbatim}
Now provide the output for the following question:
What are the stations that lie on line 7?
\end{verbatim}
}

\subsection{Evaluation} 

\begin{table}[t]
\centering
\caption{Comparison of LLMs used in our evaluation.}
\label{table:LLMComparison}
\small
\begin{tabular}{@{\extracolsep{\fill}}l@{\qquad}r@{\quad}c@{\quad}l@{\quad}l@{\quad}r}
\toprule
\textbf{Model} & \textbf{Parameters} & \textbf{Open Source} & \textbf{Price} & \textbf{Company} & \textbf{Token Limit} \\
\midrule
GPT-4 & $1.5 \times 10^{12}$ & $\times$ & USD$\,20$ p/m & OpenAI & $32\,768$\\
GPT-3.5 & $175 \times 10^{9\hphantom{1}}$ & $\times$ & free & OpenAI & $4\,096$\\
Bard & $1.6 \times 10^{12}$ & $\times$ & free & Google & $2\,048$ \\
GPT4ALL & $7 \times 10^{9\hphantom{1}}$ & \checkmark & self hosted & Nomic AI & $2\,048$ \\
Vicuna 13b & $13 \times 10^{9\hphantom{1}}$ & \checkmark & per request & Meta & $2\,048$ \\
Zephyr 7b  & $7 \times 10^{9\hphantom{1}}$ & \checkmark & free &  HuggingFace H4 & $8\,192$ \\
\bottomrule
\end{tabular}
\end{table}

\noindent
We evaluated the method from the previous section to answer to following research questions:
\begin{compactenum}
\item[(R1)] Is the method suitable for realising the language component of \ours?
\item[(R2)] What is the trade-off between grand scale LLMs and smaller, more cost-efficient alternatives?
\item[(R3)] How well does the method generalise to questions formulated in a different way than in \CLEGR?
\end{compactenum}


\paragraph{Overview of used LLMs.}

We compared different models (GPT-4, GPT3.5, Bard, GPT4All, Vicuna 13b, and Zephyr 7b; cf.~Table~\ref{table:LLMComparison})\footnote{\url{https://openai.com/research/gpt-4}; \url{https://platform.openai.com/docs/models/gpt-3-5}; \url{https://bard.google.com/}; \url{https://gpt4all.io/index.html};\url{https://huggingface.co/lmsys/vicuna-13b-v1.3}; \url{https://huggingface.co/HuggingFaceH4/zephyr-7b-beta}.}
on the semantic parsing task.

GPT-4 is the latest model developed by OpenAI with 1.5 trillion parameters and a context limit of  $32\,768$ tokens. For a price of USD$\,20$ per month, the ChatGPT Plus offer can be subscribed, allowing users to send up to 50 requests in a three-hour time frame to a hosted version of GPT-4. 

GPT3.5 is the predecessor of OpenAIs GPT-4 and is available online for free. It uses 175 billion parameters and is capable of contexts of $4\,096$ tokens.

Bard is Google's counterpart to OpenAI's dominant LLMs, using slightly more parameters than GPT-4 but has a context window of only $2\,048$ tokens. It is free to its users; however, all EU states are currently excluded from using the service due to copyright concerns.

GPT4All is an open source model that only needs 7 billion parameters. With a context limit of $2\,048$ tokens, it competes with Google Bard; however, there is no official hosted service to run this LLM. It was developed using the open source weights of Alpaca, a model developed and released by Meta. GPT-4 served as a training data generator for this model, making it a cheap alternative to expensive large-scale models.

Vicuna 13b was developed and open-sourced by Meta and comes with 13 billion parameters and a context window of $2\,018$ tokens. It serves as a middle ground between large-scale LLMs and small alternatives such as GPT4All. It is not hosted on an official server, but there are external services that host this model and even offer fine-tuning to user specific use cases.

Zephyr 7b ($\beta$) is a fine-tuned version of the Mistral 7B model. It was developed by the Hugging Face H4 team and is published under the MIT license.
    
\paragraph{Datasets.}

We created two datasets for our evaluation: \CLEGRPLUS and \CLEGRHUMAN.
The former is a straight-forward hand-crafted extension of the questions from the original \CLEGR dataset. Besides original questions that can be parsed with regular expressions, the dataset also contains versions where words are replaced with synonyms and the
position of words is slightly changed, as well as questions that entirely rephrase the original ones. For example, ``Are stations A and B on the same line?'' could be rephrased as
``Can I reach station A from station B without line change?''. The \CLEGRPLUS dataset consists of $74$ questions in total.  

\CLEGRHUMAN is a dataset that was created using an online survey. 
The survey takers were presented with a metro map and a couple of example questions.
After that, they had the task of formulating further questions such as ``Ask about the distance between Station A and Station B'' and answering their own questions. This enables cross-peer validation by having other users evaluate the same question and compare their answers.
Each surveyor had to answer a total of 12 questions; 27 people from Austria, Switzerland, and Germany aged between 18 and 33 years completed the survey, 22 of which were students.
The dataset consists of 324 questions in total.

\paragraph{Results and Discussion.}


\begin{table}[t]
\centering
\caption{Results of the evaluation on the \CLEGRPLUS dataset.}
\label{table:piechartsStandard}
\small
\begin{tabular}{@{\extracolsep{\fill}}lrrrr}
\toprule
Model       &   full match  &   contains solution   &   task missed & no answer \\
\midrule
GPT-4       &   85\%        &   0\%                 &   15\%        &   0\%     \\
GPT-3.5     &   42\%        &   8\%                 &   50\%        &   0\%     \\
Bard        &   0\%         &   76\%                &   24\%        &   0\%     \\
GPT4ALL     &   0\%         &   23\%                &   77\%        &   0\%     \\
Vicuna 13b  &   8\%         &   24\%                &   34\%        &   34\%    \\
Zephyr 7b   &   0\%         &   61\%                &   21\%        &   18\%    \\
\bottomrule
\end{tabular}

\end{table}




\begin{table}[t]
\centering
\caption{Results of the evaluation on the \CLEGRHUMAN dataset.}
\label{table:piechartsHuman}
\small
\begin{tabular}{@{\extracolsep{\fill}}lrrrr}
\toprule
Model       &   full match  &   contains solution   &   task missed & no answer \\
\midrule
GPT-4       &   94\%        &   0\%                 &   6\%        &   0\%     \\
GPT-3.5     &   38\%        &   7\%                 &   55\%       &   0\%     \\
Bard        &   0\%         &   78\%                &   22\%       &   0\%     \\
GPT4ALL     &   0\%         &   16\%                &   84\%       &   0\%     \\
Vicuna 13b  &   4\%         &   19\%                &   46\%        &  31\%    \\
Zephyr 7b   &   0\%         &   72\%                &   17\%        &  11\%    \\
\bottomrule
\end{tabular}

\end{table}


The results of our evaluation are summarised in Table~\ref{table:piechartsStandard} for
\CLEGRPLUS and in Table~\ref{table:piechartsHuman} for \CLEGRHUMAN.
We classified the answers produced by the LLMs into four categories:
\begin{compactitem}

\item {\emph{full match}}: 
the response matches exactly the set of expected atoms;

\item {\emph{contains solution}}: 
the expected atoms can be extracted from the respone;

\item {\emph{task missed}}: 
the response contains some text but not the expected atoms; 

\item {\emph{no answer}}: 
the response consists of only whitespace characters.
\end{compactitem}

Note that ``full match'' as well as ``contains solution'' can be used for the ASP reasoning task, while answers from the other categories cannot be used.

GPT-4 performed best among the considered LLMs as it produced $85\%$ completely correct responses on \CLEGRPLUS and even 94\% on \CLEGRHUMAN. It also always provided an answer to the prompt. 
GPT-3.5 invented new predicates for half of the questions. For the remaining ones, its response matched exactly or contained the solution.
Vicuna often did not give a proper response due to context overflow and trails behind GPT-4 and GPT-3.5 also in terms of correct answers.
Google Bard never got the exact solution due to extensive additional explanations for all predicates no matter the prompt. However, the responses contained the solution in about three quarters of the cases. In this regard, it is only outmatched by GPT-4.  
This performance is similar to that of the much smaller open-source model Zephyr 7b, which is trailing only slightly behind.
The responses of GPT4All contain the correct solution for only $23\%$ (\CLEGRPLUS) and $16\%$ (\CLEGRHUMAN) of the questions.

We answer our initial research questions therefore as follows:
At least GPT-4 is suitable for realising the language component with an acceptable trade-off between accuracy and ability to generalise (R1).
Although GPT-4 exhibits the best overall performance, especially the free and much smaller Zephyr model shows promising results (R2). Throughout, the LLMs perform similarly on \CLEGRPLUS and \CLEGRHUMAN, which showcases the strength of LLMs for language processing without the need for context-specific training (R3).

\section{Related Work}\label{sec:rel}

Our approach 
builds on previous work~\cite{eiterHOP22}, where we introduced a 
neuro-symbolic method for VQA in the context of the CLEVR dataset~\cite{johnson2017clevr}
using a reasoning component based on ASP inspired by NSVQA~\cite{yi2018nsvqa}. The latter used a combination of RCNN~\cite{ren2015rcnn} for object detection, an LSTM~\cite{hochreiter1997lstm} for natural language parsing, and Python as a symbolic executor to infer the answer. 
The vision and language modules in these previous approaches were trained for the datasets. 
As compared to these datasets the number of questions obtained from the questionnaires to build our dataset is small, it would be hardly possible to effectively train an 
LSTM on them. 
It is a particular strength of our work that we resort to LLMs that do not require any further training.

We also mention the neural and end-to-end trainable MAC system~\cite{hudson2018mac} that achieves very promising results in VQA datasets, provided there is enough data available to train the system.
A recent approach that combines large pretrained models for images and text in combination with symbolic execution in Python is ViperGPT~\cite{vipergpt}; complicated graph images
are not handled well by pretrained vision-language models, however.

A characteristic of \ours is that we use ASP for reasoning, an
idea that was also explored in previous work~\cite{RileyS19,BasuSG20,eiterHOP22,EiterGHO23}.
Outside of the context of VQA, ASP has been applied for various neuro-symbolic tasks such as
segmentation of laryngeal images~\cite{brunoCMM21}, and discovery of rules that explain sequences of sensory input~\cite{evansHWKS21}. 
Barbara et al.~\cite{BarbaraGLMQRR23} describe a neuro-symbolic approach that involves ASP for visual validation of electric panels where a component graph from an image is matched against its specification. This is an example of another interesting application that involves images of graphs and our approach could be used to contribute question-answering capabilities in such a setting.

In passing, it should be noted that there are also systems that can be used for neuro-symbolic learning, \egc by employing semantic loss~\cite{xu2018sl}, which means that they use the information produced by the reasoning module to improve the learning tasks of the neural networks involved~\cite{yang2020neurasp,ManhaeveDKDR21}.

Our approach to using LLMs to extract predicates for the downstream reasoning task is inspired by recent work by Rajasekharan et al.~\cite{rajasekharan23}. They proposed the STAR framework, which consists of LLMs and prompts for extracting logical predicates in combination with an ASP knowledge base. The authors applied STAR to different problems requiring qualitative reasoning, mathematical reasoning, as well as goal-directed conversation. 
Going one step further, Ishay et al.~\cite{ishay23} introduced a method to translate problems formulated in natural language into complete ASP programs. This method requires multiple prompts, each responsible for a subtask such as identifying constant symbols, forming predicates, and transforming the specification into rules. 
The idea to apply LLMs to parse natural language into a formal language suitable for automated reasoning is also found outside the context of ASP, \egc work by Liu et al.~\cite{liu23}, who use prompting techniques to translate text into the Planning Domain Definition Language.

\section{Conclusion}\label{sec:concl}

We addressed the relevant the problem of integrating ASP with vision and language modules to solve a new VQA variant that is concerned with images of graphs.
For this task, we introduced a respective dataset that is based on an existing one for graph question answering on transit networks, and we presented \ours, a modular neuro-symbolic model for VGQA that combines neural components for graph and question parsing and symbolic reasoning with ASP for question answering.  We studied LLMs for the question parsing component to improve how well our method generalises to unseen questions. \ours has been evaluated on the VGQA dataset and therefore constitutes a first baseline for the novel dataset. 

The advantages of a modular architecture in combination with logic programming are that
the solution is transparent, interpretable, explainable, easier to debug, and components can be replaced with better ones over time in contrast to more monolithic end-to-end trained models. 
Our system notably relies on pretrained components and thus requires no additional training. With the advent of large pretrained models for language and images such as GPT-4~\cite{openai2023gpt4} or CLIP~\cite{RadfordKHRGASAM21},
such architectures, where symbolic systems are used to control and connect neural ones, may be seen more frequently.

For future work, we plan to look into better alternatives for the visual module that is more suitable for complicated images of graphs, which is currently the limiting factor. 
Another future direction is to work with real-world metro networks for which currently no VQA datasets exist.

\bibliographystyle{eptcs}
\bibliography{refs}
\end{document}